%% file: sn-article.tex
\documentclass[pdflatex,sn-mathphys-num]{sn-jnl}

\usepackage{graphicx}%
\usepackage{multirow}%
\usepackage{amsmath,amssymb,amsfonts}%
\usepackage{amsthm}%
\usepackage{mathrsfs}%
\usepackage[title]{appendix}%
\usepackage{xcolor}%
\usepackage{textcomp}%
\usepackage{manyfoot}%
\usepackage{booktabs}%
\usepackage{algorithm}%
\usepackage{algorithmicx}%
\usepackage{algpseudocode}%
\usepackage{listings}%
\usepackage[T1]{fontenc}
\usepackage{caption}

\begin{document}

\title[Article Title]{Reducing Large Language Model Safety Risks in Women's Health using Semantic Entropy}


\author[1]{\fnm{Jahan C.} \sur{Penny-Dimri}}

\author[2]{\fnm{Magdalena} \sur{Bachmann}}

\author[1]{\fnm{William R.} \sur{Cooke}}

\author[2]{\fnm{Sam} \sur{Mathewlynn}}

\author[3]{\fnm{Samuel} \sur{Dockree}}

\author[1]{\fnm{John} \sur{Tolladay}}

\author[4]{\fnm{Jannik} \sur{Kossen}}

\author[4]{\fnm{Lin} \sur{Li}}

\author[4]{\fnm{Yarin} \sur{Gal}}

\author*[1]{\fnm{Gabriel Davis} \sur{Jones}}\email{gabriel.jones@wrh.ox.ac.uk}

\affil[1]{\orgdiv{Oxford Digital Health Labs, Nuffield Department of Women's and Reproductive Health}, \orgname{University of Oxford}, \orgaddress{\city{Oxford}, \country{UK}}}

\affil[2]{\orgdiv{Nuffield Department of Women's and Reproductive Health}, \orgname{University of Oxford}, \orgaddress{\city{Oxford}, \country{UK}}}

\affil[3]{\orgname{Oxford University Hospitals NHS Foundation Trust}, \orgaddress{\city{Oxford}, \country{UK}}}

\affil[4]{\orgdiv{OATML, Department of Computer Science}, \orgname{University of Oxford}, \orgaddress{\city{Oxford}, \country{UK}}}


\abstract{Large language models (LLMs) hold substantial promise for clinical decision support. However, their widespread adoption in medicine, particularly in healthcare, is hindered by their propensity to generate false or misleading outputs, known as hallucinations. In high-stakes domains such as women's health (obstetrics \& gynaecology), where errors in clinical reasoning can have profound consequences for maternal and neonatal outcomes, ensuring the reliability of AI-generated responses is critical. Traditional methods for quantifying uncertainty, such as perplexity, fail to capture meaning-level inconsistencies that lead to misinformation. Here, we evaluate semantic entropy (SE), a novel uncertainty metric that assesses meaning-level variation, to detect hallucinations in AI-generated medical content. Using a clinically validated dataset derived from UK RCOG MRCOG examinations, we compared SE with perplexity in identifying uncertain responses. SE demonstrated superior performance, achieving an AUROC of 0.76 (95\% CI: 0.75–0.78), compared to 0.62 (0.60–0.65) for perplexity. Clinical expert validation further confirmed its effectiveness, with SE achieving near-perfect uncertainty discrimination (AUROC: 0.97). While semantic clustering was successful in only 30\% of cases, SE remains a valuable tool for improving AI safety in women's health. These findings suggest that SE could enable more reliable AI integration into clinical practice, particularly in resource-limited settings where LLMs could augment care. This study highlights the potential of SE as a key safeguard in the responsible deployment of AI-driven tools in women's health, leading to safer and more effective digital health interventions.}



\maketitle

\input{sections/introduction}

\input{sections/results}

\input{sections/methods}

\input{sections/discussion}

\input{sections/conclusion}


\section*{Supplementary information}

Supplementary tables can be found at \href{https://tinyurl.com/r3znc99e}{https://tinyurl.com/r3znc99e}.

\section*{Declarations}
The authors declare no conflicts of interest.




\end{document}

%% file: sections/introduction.tex
\section{Introduction}\label{sec1}

Large language models (LLMs) have transformed how information is processed and applied across various fields. Advanced LLMs like ChatGPT have demonstrated capabilities that surpass human performance in some benchmarks of clinical knowledge~\cite{Bachmann2024}. By learning from vast amounts of data, these models can generate responses that mimic human language fluency, making them appealing tools for enhancing healthcare. In clinical settings, LLMs are theorised to expedite decision-making by providing rapid access to medical knowledge, potentially reducing diagnostic delays and improving care quality~\cite{Singhal2023}. 

In women’s health, particularly obstetrics and gynaecology (O\&G), LLMs hold promise in addressing long-standing critical gaps in diagnosis and treatment~\cite{Gronowski2018, Clancy1992, Owens2008}. The potential of these models is particularly compelling for resource-limited settings, where they could help bridge gaps in healthcare delivery. O\&G has long been characterised by diagnostic and treatment gaps globally. These disproportionately impact maternal and neonatal outcomes, exacerbating health inequities~\cite{Amin2021, Peneva2015}. Accurate diagnosis and timely management are essential in this domain, as delays or errors can have severe, life-altering consequences. Safely developed LLMs could transform care delivery by providing reliable, evidence-based insights to those in resource-limited settings where expertise is scarce~\cite{Cascella2023, Antaki2023}. However, their integration into clinical practice must address critical concerns about reliability, safety, and the propagation of misinformation, as the consequences of medical decision-making demand rigorous validation and oversight~\cite{Bachmann2024, Wang2023}.

A critical barrier to LLM adoption in clinical contexts is their tendency to produce ``hallucinations''—responses that appear coherent but are factually incorrect or ungrounded~\cite{Kocon2023, Temsah2023}. Hallucinations pose a particular risk in healthcare, where misinformation can lead to adverse outcomes for patients. This issue is exacerbated when models encounter questions requiring nuanced clinical reasoning or domain-specific expertise~\cite{Kung2023}. Despite efforts to refine LLM performance, hallucinations remain a pervasive problem, undermining trust in these technologies. Addressing this limitation is essential for advancing their utility in this high-stakes environment, where accuracy and reliability are paramount.

One promising strategy to mitigate hallucinations is improving how uncertainty is measured and therefore mitigated within LLM-generated responses. Traditional uncertainty quantification methods, which often rely on token-level variations, struggle to capture inconsistencies in meaning~\cite{Kuhn2023}. Semantic entropy, a recently developed metric, provides a more robust framework by assessing uncertainty at the level of meaning rather than individual words~\cite{Kuhn2023}. Unlike conventional approaches that focus on lexical variation, semantic entropy quantifies uncertainty across clusters of semantically equivalent responses, enabling the detection of confabulations—LLM outputs that are both erroneous and arbitrary. This method represents an important step toward improving the reliability of LLM-generated content, particularly in free-form text generation tasks, where traditional uncertainty measures often fall short.

This study applies semantic entropy to evaluate the performance of ChatGPT on a clinically-validated dataset derived from the UK Royal College of Obstetricians and Gynaecologists (RCOG) MRCOG Part One and Part Two examinations. These exams serve as rigorous international benchmarks for assessing specialist clinical knowledge and reasoning in O\&G, providing an ideal context to test the efficacy of semantic entropy in detecting hallucinations. Crucially, this dataset is not in the public domain, ensuring it has not contributed to the development of any current LLMs~\cite{Bachmann2024}. This ensures the performance of LLMs is tested on unseen data, offering insights into the practical applicability of semantic entropy for advancing safety and accuracy in women’s health. By analysing 1,824 examination questions covering both foundational and applied clinical knowledge, this study evaluates the effectiveness of semantic entropy in assessing LLM reliability. GPT-4o’s responses undergo additional expert clinical validation, providing insights into the practical utility of semantic entropy in improving the safety and accuracy of LLM-generated medical content in women’s health.

%% file: sections/results.tex
\section{Results}\label{sec2}

1,824  MRCOG questions were compiled from eight distinct sources and reformatted for compatibility with GPT-4o. Each question was reviewed by certified clinical experts (specialists) in O\&G. The dataset included 835 Part One questions and 989 Part Two questions, categorised into knowledge domains defined by the RCOG: 14 domains for Part One and 15 for Part Two. The median number of questions per domain was 58 (IQR 32--85) for Part One and 45 (IQR 32--64) for Part Two. Filtering excluded 126 questions incompatible with short-answer formats and 54 questions requiring interpretation of images or tables, resulting in a final dataset of 1,644 questions: 780 from Part One and 864 from Part Two. Of these, 590 questions assessed clinical reasoning, while 1,080 tested factual knowledge.

\subsection{Semantic Entropy Outperforms Perplexity in Measuring Uncertainty}
Semantic entropy (SE), its discrete version, and perplexity were evaluated as metrics of uncertainty using accuracy and area under the receiver operating characteristic curve (AUROC) \cite{Farquhar2024}.  Accuracy was defined as the percentage of correct answers. Accuracy was expected to remain constant across different uncertainty metrics, as it is solely determined by whether a response matches the correct answer, i.e. it is independent of how uncertainty is measured. Uncertainty metrics, such as semantic entropy and perplexity, assess the model’s confidence in its responses but do not alter the correctness of those responses. The AUROC quantifies how well a metric distinguishes between correct and incorrect answers, regardless of thresholds, making it a standard measure of response uncertainty. SE and its discrete variant significantly outperformed perplexity in uncertainty discrimination, achieving AUROCs of 0.76 (0.75 - 0.78) and 0.75 (0.73 - 0.78), respectively, compared to 0.62 (0.60 - 0.65) for perplexity.  Accuracy was consistent across metrics (50\%), where the lowest perplexity response and largest semantic cluster showed statistically similar performance. (Table~\ref{tab:baseline_performance}).

\begin{table}[h!]
\centering
\caption{\textbf{Performance of Semantic Entropy and Perplexity in Uncertainty Discrimination Across the MRCOG Dataset.} Semantic entropy and its discrete variant demonstrate higher uncertainty discrimination than perplexity when tested on MRCOG examination questions, with model temperature set at 1.0. Higher AUROC values indicate that semantic entropy more effectively differentiates between correct and incorrect responses. Accuracy remains similar across all methods, confirming that uncertainty metrics influence confidence estimation rather than correctness.}
\label{tab:baseline_performance}
\begin{tabular}{@{}lcc@{}}
\toprule
\textbf{Metric}    & \textbf{Accuracy (95\% CI)}  & \textbf{AUROC (95\% CI)}      \\
\midrule
Semantic Entropy (SE)            & 0.50 (0.48 -- 0.52)          & 0.76 (0.73 -- 0.78)          \\
Discrete SE & 0.50 (0.48 -- 0.52)          & 0.75 (0.73 -- 0.78)          \\
Perplexity         & 0.51 (0.48 -- 0.53)          & 0.62 (0.60 -- 0.65)          \\
\bottomrule
\end{tabular}
\end{table}

\subsection{Semantic Entropy Outperforms Perplexity in Knowledge and Reasoning Tasks}
\begin{table}[ht]
\centering
\caption{\textbf{Comparative Performance of Semantic Entropy and Perplexity Across MRCOG Part One and Part Two Questions.} Semantic entropy and its discrete variant exhibit higher uncertainty discrimination than perplexity for both factual knowledge (Part One) and clinical reasoning (Part Two) questions. Accuracy and AUROC values are higher for Part One, indicating that knowledge retrieval tasks yield better-calibrated uncertainty measures. Part Two questions introduce greater variability due to their reasoning-based format, leading to lower accuracy and increased model uncertainty. Semantic entropy maintains a consistent advantage in AUROC across both parts, demonstrating improved uncertainty estimation compared to perplexity.}
\label{tab:part_subgroup}
\begin{tabular}{llcc}
\toprule
\textbf{Metric}     & \textbf{Part} & \textbf{Accuracy (95\% CI)}  & \textbf{AUROC (95\% CI)}  \\
\midrule
Semantic Entropy (SE)     & Part 1 & 0.58 (0.54 -- 0.61) & 0.77 (0.73 -- 0.80) \\
            & Part 2 & 0.43 (0.40 -- 0.46) & 0.73 (0.70 -- 0.76) \\
Discrete SE    & Part 1 & 0.58 (0.54 -- 0.61) & 0.75 (0.72 -- 0.79) \\
            & Part 2 & 0.43 (0.40 -- 0.46) & 0.73 (0.70 -- 0.77) \\
Perplexity  & Part 1 & 0.60 (0.57 -- 0.64) & 0.66 (0.62 -- 0.70) \\
            & Part 2 & 0.42 (0.39 -- 0.46) & 0.59 (0.55 -- 0.63) \\
\bottomrule
\end{tabular}
\end{table}
Subgroup analyses revealed performance differences between Part One and Part Two questions. Questions in Part One generally focus on knowledge retrieval whereas Part Two tests clinical reasoning.
There is, however, significant overlap in the types of questions between Parts. We therefore explored the difference in performance across questions labelled as knowledge retrieval versus reasoning by the LLM. The LLM scored statistically significantly higher accuracy on Part One questions and trended toward better uncertainty calibration, with a higher AUROC on Part One questions (Table~\ref{tab:part_subgroup}).
SE had better uncertainty calibration compared to perplexity across Part One, 0.77 (0.73 - 0.80) vs 0.66 (0.62 - 0.70), and Part Two, 0.73 (0.70 - 0.76) vs 0.59 (0.55 - 0.63).
Similarly, questions classified as knowledge retrieval had statistically significantly higher accuracy, consistent with trends observed in Part One and Part Two comparisons. (Table~\ref{tab:reasoning_subgroup}).
While the trend in the AUROC was reversed, with SE slightly outperforming in reasoning tasks, these results did not achieve statistical significance.
Importantly, SE had better uncertainty calibration than perplexity across both knowledge tasks, 0.74 (0.72 - 0.77) vs 0.67 (0.64 - 0.70), and reasoning tasks, 0.77 (0.73 – 0.81) vs 0.66 (0.61 – 0.70).

\begin{table}[ht]
\centering
\caption{\textbf{Accuracy and Uncertainty Discrimination in Knowledge vs. Reasoning Tasks.} Accuracy and uncertainty discrimination of semantic entropy, its discrete variant, and perplexity were evaluated for knowledge retrieval and clinical reasoning tasks. Semantic entropy achieves higher AUROC than perplexity across both task types, indicating better uncertainty calibration. Accuracy is lower for reasoning tasks, reflecting increased model uncertainty and greater variability in generated responses. The AUROC advantage is more pronounced in reasoning tasks, suggesting improved robustness in detecting uncertainty in complex clinical decision-making scenarios.}
\label{tab:reasoning_subgroup}
\begin{tabular}{llcc}
\toprule
\textbf{Metric}      & \textbf{Category} & \textbf{Accuracy (95\% CI)}  & \textbf{AUROC (95\% CI)} \\
\midrule
Semantic Entropy (SE)     & Knowledge  & 0.56 (0.53 -- 0.59) & 0.74 (0.72 -- 0.77) \\
            & Reasoning  & 0.39 (0.35 -- 0.43) & 0.77 (0.73 -- 0.81) \\
Discrete SE    & Knowledge  & 0.56 (0.53 -- 0.59) & 0.74 (0.71 -- 0.76) \\
            & Reasoning  & 0.39 (0.35 -- 0.43) & 0.76 (0.72 -- 0.80) \\
Perplexity  & Knowledge  & 0.58 (0.55 -- 0.60) & 0.67 (0.64 -- 0.70) \\
            & Reasoning  & 0.38 (0.34 -- 0.42) & 0.66 (0.61 -- 0.70) \\
\bottomrule
\end{tabular}
\end{table}

Shorter response sequences achieved significantly higher accuracy and AUROC across all metrics compared to longer responses, reflecting better correctness and uncertainty discrimination (Table~\ref{tab:length_subgroup}). SE demonstrated better uncertainty discrimination compared to perplexity on long responses with an AUROC of 0.73 (0.69 - 0.78) compared with 0.64 (0.59 - 0.68).

\begin{table}[ht]
\centering
\caption{\textbf{Effect of Response Length on Accuracy and Uncertainty Discrimination.} The accuracy and uncertainty discrimination of semantic entropy, its discrete variant, and perplexity were assessed for short (\textless 15 characters) and long (\textgreater 60 characters) AI-generated responses. Short responses achieve higher accuracy and AUROC across all metrics, reflecting greater model confidence and reliability in concise outputs. SE outperforms perplexity for both short and long responses, but longer outputs introduce greater semantic variability, reducing overall accuracy and making uncertainty estimation less precise.}
\label{tab:length_subgroup}
\begin{tabular}{llcc}
\toprule
\textbf{Metric}     & \textbf{Length} & \textbf{Accuracy (95\% CI)}  & \textbf{AUROC (95\% CI)}  \\
\midrule
Semantic Entropy (SE)     & Short & 0.88 (0.73 -- 0.95) & 0.88 (0.74 -- 1.00) \\
            & Long  & 0.36 (0.33 -- 0.41) & 0.73 (0.69 -- 0.78) \\
Discrete SE    & Short & 0.88 (0.73 -- 0.95) & 0.79 (0.59 -- 0.99) \\
            & Long  & 0.36 (0.33 -- 0.41) & 0.73 (0.68 -- 0.77) \\
Perplexity  & Short & 0.82 (0.66 -- 0.91) & 0.82 (0.66 -- 0.98) \\
            & Long  & 0.38 (0.35 -- 0.43) & 0.64 (0.59 -- 0.68) \\
\bottomrule
\end{tabular}
\end{table}

\subsection{Higher Temperature Improves Uncertainty Discrimination}
\begin{table}[ht]
\centering
\caption{\textbf{Effect of Temperature on Accuracy and Uncertainty Discrimination in AI-Generated Responses.} The impact of model temperature (0.2 vs. 1.0) on uncertainty estimation was assessed using semantic entropy, its discrete variant, and perplexity. Accuracy (95\% CI) represents the proportion of correct responses, while AUROC (95\% CI) quantifies the ability of each metric to distinguish between correct and incorrect answers. Increasing temperature from 0.2 to 1.0 leads to higher AUROC values across all uncertainty metrics, indicating improved uncertainty discrimination at greater response variability. SE maintains a higher AUROC than perplexity at both temperature settings, suggesting better calibration of model confidence. Accuracy remains stable across conditions, confirming that temperature primarily affects uncertainty estimation rather than correctness.}
\label{tab:temperature_subgroup}
\begin{tabular}{llcc}
\toprule
\textbf{Metric} & \textbf{Temp} & \textbf{Accuracy (95\% CI)}  & \textbf{AUROC (95\% CI)} \\
\midrule
Semantic Entropy (SE)   & 0.2 & 0.51 (0.48 -- 0.53) & 0.71 (0.68 -- 0.73) \\
          & 1.0 & 0.50 (0.48 -- 0.52) & 0.76 (0.73 -- 0.78) \\
Discrete SE  & 0.2 & 0.51 (0.48 -- 0.53) & 0.67 (0.65 -- 0.70) \\
          & 1.0 & 0.50 (0.48 -- 0.52) & 0.75 (0.73 -- 0.78) \\
Perplexity & 0.2 & 0.52 (0.50 -- 0.55) & 0.58 (0.55 -- 0.61) \\
           & 1.0 & 0.51 (0.48 -- 0.53) & 0.62 (0.60 -- 0.65) \\
\bottomrule
\end{tabular}
\end{table}
The effect of temperature on uncertainty metrics was assessed by comparing performance at temperatures of 0.2 and 1.0. AUROC increased for all metrics as temperature rose, indicating improved discrimination of uncertainty at higher randomness levels (Table~\ref{tab:temperature_subgroup}). As expected, accuracy was stable across both temperatures, with SE and its discrete variant maintaining similar performance. 

\subsection{Clinical Expert Validation}
\begin{table}[ht]
\centering
\caption{\textbf{Accuracy of AI-Generated Responses Evaluated by Clinical Experts and the LLM, Stratified by Semantic Clustering Method.} Three Clinical experts evaluated the correctness of AI-generated responses to a subset of MRCOG questions. Responses were grouped into semantic clusters using semantic entropy, where a lower cluster count indicates greater consistency in model outputs. Accuracy is reported for responses selected using two methods: (i) Lowest Perplexity, where the model-selected response has the lowest perplexity score, and (ii) Largest Cluster, where the most frequently generated meaning-based response is chosen. For each selection method, accuracy was assessed in two ways: Clinical Expert Scored, where O\&G specialists determined correctness, and LLM Scored, where correctness was assessed based on bidirectional entailment with the reference answer. Accuracy was highest when responses formed a single cluster, while increasing cluster count corresponded to greater uncertainty. Expert validation indicates that semantic clustering was fully successful in only 30\% of cases but remains informative for uncertainty estimation.}
\label{tab:human_llm_comparison}
\begin{tabular}{ccccc}
\toprule
\textbf{} & \multicolumn{2}{c}{\textbf{Lowest Perplexity}} & \multicolumn{2}{c}{\textbf{Largest Cluster}} \\
\cmidrule(lr){2-3} \cmidrule(lr){4-5}
\textbf{Clusters} & \textbf{Clinical Expert Scored}& \textbf{LLM Scored} & \textbf{Clinical Expert Scored }& \textbf{LLM Scored} \\
\midrule
1 & 57.14\% & 85.71\% & 90.48\% & 85.71\% \\
2 & 36.84\% & 42.11\% & 10.53\% & 31.58\% \\
3 & 50.00\% & 25.00\% & 0.00\% & 25.00\% \\
4 & 13.33\% & 6.67\% & 0.00\% & 6.67\% \\
5 & 10.53\% & 0.00\% & 5.26\% & 10.53\% \\
6 & 50.00\% & 12.50\% & 0.00\% & 0.00\% \\
7 & 50.00\% & 0.00\% & 0.00\% & 50.00\% \\
8 & 33.33\% & 0.00\% & 0.00\% & 0.00\% \\
\bottomrule
\end{tabular}
\end{table}
Three  O\&G specialists evaluated a set of 105 randomly selected MRCOG questions and responses from ChatGPT. A strong relationship between semantic clustering and response accuracy was observed. Consistent with expectations, single-cluster responses achieved the highest accuracy, 90.48\% , while accuracy decreased with increasing number of clusters. Semantic clustering was successful for only 30\% of questions, where success was defined as all clusters having a unique meaning and all responses within a cluster having the same meaning. Despite the high error rate in clustering, responses grouped by meaning were effective for uncertainty analysis. These findings underscore the potential of SE for improving LLM reliability in clinical applications (Table~\ref{tab:human_llm_comparison}).
The results from the clinical expert validation are shown in Table S5 of the Online Supplementary Material.

\subsection{Definition of Correctness does not Affect Uncertainty Calibration}

We tested how the definition of a correct response affected uncertainty discrimination results when correctness was assessed by the LLM. 
We tested four definitions of correctness, and while different definitions significantly affected the accuracy of the model, they did not statistically significantly affect the uncertainty discrimination of the semantic entropy metric. These results are shown in Tables S1, S2, S3, and S4 in the Online Supplementary Material.

%% file: sections/methods.tex
\section{Methods}\label{sec11}

\subsection{Data Source and Processing}
Domain-specific questions were adapted from single best answer (SBA) and extended matching questions (EMQ) from the MRCOG Part One and Part Two examinations. These questions were sourced from a private database inaccessible to publicly available LLMs and restricted to items created after 2015 to ensure alignment with contemporary clinical practice. This  dataset has previously been described~\cite{Bachmann2024}.

Questions first underwent a rigorous preprocessing pipeline. Each question-answer pair was validated through a dual-review process to ensure both technical accuracy of format conversion and clinical relevance \cite{Bachmann2024}. Questions requiring contextual information were augmented with necessary details, while those containing repeated content or requiring image analysis were excluded \cite{Bachmann2024}. All questions were rephrased to conform to a short-answer format. The model answer for each question was derived from the correct option within the original SBA or EMQ.

\subsection{Inference Settings, Prompt Engineering, and Response Generation}
Inference experiments were conducted using the frontier OpenAI model, GPT-4o (gpt-4o-2024-08-06), accessed via their application programming interface (API) \cite{openAIAPI}. Prompts were designed following established best practices and are available on the published codebase \cite{OpenAIPrompt, White2023}.

Output randomness was controlled using the temperature parameter, where a value of 0.0 produces deterministic responses \cite{openAIAPI}. For generating responses, the temperature was set at 1.0. A sensitivity analysis was performed with a lower temperature of 0.2 to examine variability in responses. For each question 10 responses were generated. 

\subsection{Measuring Uncertainty and Semantic Entropy}
Two metrics were used to quantify uncertainty: perplexity and semantic entropy (SE). Perplexity, a standard token-level metric, aggregates token confidence as the exponentiation of the average negative log-likelihood of a sequence. Lower perplexity indicates higher model confidence.

Semantic entropy, a recently developed metric \cite{Kuhn2023, Farquhar2024}, evaluates uncertainty at the semantic level. Unlike perplexity, SE measures variability in meaning across multiple generated responses. The computation of SE involves: (1) generating a set of \(M\) responses for a given prompt, (2) clustering these responses based on semantic similarity using bidirectional entailment, and (3) calculating entropy based on the distribution of responses across clusters. Lower SE indicates greater confidence in the model’s responses. Additionally, a discrete version of SE, calculated without access to token-level log-probabilities, was included \cite{Farquhar2024}. To assess bidirectional entailment, the temperature was fixed at 0.0 to ensure deterministic behaviour.

\subsubsection{Semantic Clustering Procedure}
Semantic clustering followed established protocols \cite{Kuhn2023, Farquhar2024}. Responses were iteratively grouped into clusters if they shared semantic meaning, as determined by bidirectional entailment \cite{Pado2009}. For example, the statements \textit{``Preeclampsia is characterised by hypertension and proteinuria after 20 weeks of gestation''} and \textit{``Hypertension and proteinuria occurring after 20 weeks indicate preeclampsia''} are considered semantically equivalent due to their shared meaning.

\subsection{Correctness of Responses}
Correctness was defined by the bidirectional entailment between a model’s response and the reference answer. Two correctness criteria were applied, depending on the uncertainty metric:

\begin{enumerate}
    \item For perplexity, the response with the lowest perplexity was deemed correct if it was bidirectionally entailed by the reference answer.
    \item For SE, the largest semantic cluster represented the highest-confidence meaning.
\end{enumerate}

A response was correct if the lowest perplexity response within this cluster was bidirectionally entailed by the reference answer. If two or more clusters were equally large, the response was considered incorrect.

Sensitivity analyses evaluated alternative definitions of correctness for SE: (1) strict, where all responses in the largest cluster had to be entailed by the reference answer; (2) majority vote, where more than 50\% of responses in the largest cluster were required to be entailed; and (3) relaxed, where any response in the largest cluster needed to be entailed.

\subsection{Clinical Expert Validation}
A subset of 105 questions, along with their generated responses and clustering results, was randomly selected for human clinical expert validation. Three certified O\&G specialists independently assessed the questions, response correctness, and the clustering of responses by meaning.
The clinician was presented with the question, the true correct answer, the lowest perplexity answer, and the generated responses clustered by meaning. Feedback was obtained assessing the quality of the question, whether the lowest perplexity answer was the same as the true answer, whether the lowest perplexity answer was correct but different from the true answer, whether each cluster had a consistent and distinct meaning, and whether each cluster's meaning was equivalent to the true answer.

The findings from the clinical expert validation subset were compared against results from the automated dataset to ensure consistency and identify discrepancies.

\subsubsection{Subgroup Analysis}
Performance was stratified across subgroups to explore variability in metrics. Comparisons included Part One versus Part Two MRCOG questions, which predominantly assess knowledge retrieval and clinical reasoning, respectively. Additionally, performance was analysed by question type (knowledge versus reasoning tasks) and response length.

To examine the effect of response length, questions were classified as short (\textless15 characters) or long (\textgreater60 characters), excluding mid-length responses (16–59 characters). This approach ensured a clear contrast between distinct length-based groups. Binarising at the median or mean would result in largely similar samples, limiting meaningful comparisons. By focusing on the tails of the distribution, we aimed to better assess how response length influences model performance and uncertainty calibration.

\subsection{Statistical Analysis}
Question-answering accuracy was measured as the proportion of correct responses. AUROC was used to assess how well uncertainty metrics distinguished correct from incorrect responses. A score of 0.5 indicated no correlation, while 1.0 represented perfect correlation. 95\% confidence intervals were calculated for the AUROC.

\subsection{Codebase}
The code for our analysis is publically available for review \cite{codebaseRepo}.

%% file: sections/discussion.tex
\section{Discussion}\label{sec12}

This study introduces semantic entropy, the novel measure of model uncertainty, to enhance the reliability and safety of LLM responses in women's health. We evaluated SE using a large, private dataset of MRCOG examination questions and benchmarked its performance against a standard metric, perplexity. 

Previous work on this dataset confirmed a tendency of LLMs to generate incorrect but confident responses in this domain \cite{Bachmann2024}. Our findings extend these prior findings by demonstrating SE significantly outperforms perplexity in discriminating model uncertainty across both knowledge retrieval and clinical reasoning tasks. These results align with earlier studies describing SE's utility in detecting hallucinations and improving model responses through fine-tuning \cite{Kuhn2023, Farquhar2024}.

We also assessed the effect of response length on SE's performance, as longer responses offer more opportunities for varied expressions of equivalent meanings, potentially affecting discrimination accuracy. While we observed a trend toward reduced uncertainty discrimination with increasing response length, this effect was not significant. These findings affirm the robustness of SE across different response lengths, though our dataset primarily consisted of short-answer questions, limiting conclusions about longer responses.

Clinical expert validation further confirmed SE's ability to improve uncertainty discrimination. When correctness of answers were evaluated by O\&G specialists, SE demonstrated near-perfect discrimination, with an AUROC of 0.97 (95\% CI: 0.91–1.00), compared to 0.57 (95\% CI: 0.45–0.68) for perplexity. Notably, the human validation process reclassified correctness labels based on domain expertise. The effectiveness of SE was under-estimated when correctness of responses was scored by the LLM itself, suggesting LLMs underperform in entailment tasks involving generated responses and ground-truth answers. This implies that the effects observed in our primary analysis may underestimate SE's true potential. Human validation also revealed that the semantic clustering process was imperfect, achieving success in only 30\% of cases. Despite our strict definition of success, SE's empirical performance indicates that semantic clusters are still valuable for measuring uncertainty. Domain-specific LLMs or models explicitly designed to capture semantics, such as large concept models, could further improve the clustering process and SE's efficacy \cite{lcm2024}.

AI safety remains a critical global concern \cite{Junior2024, Bedi2024, Salvagno2023}. Preventing unpredictable problems, such as hallucinations, is essential to ensuring safe deployment of these systems \cite{Farquhar2024, Pal2023}. Women's health poses unique challenges, particularly in O\&G, where misinformation can have severe consequences. The heightened risks in this domain underscore the necessity of validated, reliable models \cite{Eoh2024, Sengupta2023}. Misinformation often arises from inadequate training data and limited domain specificity, resulting in biases that can exacerbate gender disparities \cite{Chen2022, Levy2024}. Without systematic model validation, deploying LLMs risks amplifying these inequities. Our findings confirm prior observations that GPT-4 underperforms in clinical reasoning tasks. However, we demonstrate that SE is a valuable tool for identifying true uncertainty, enabling LLMs to filter uncertain responses and enhance safety.

SE also has the potential to mitigate model bias. Addressing bias in LLM systems requires continuous review of data and outputs. While human expert validation remains the gold standard, it is resource-intensive and subject to inter-clinician variability \cite{Shankar2024}. By contrast, SE offers a scalable, probabilistic framework for performance monitoring. Measuring uncertainty can help identify demographic, socioeconomic, or cultural biases, which can then be corrected through data augmentation and domain-specific feedback loops. These strategies, facilitated by SE, could promote fairer, more representative model performance \cite{Ziegler2019, Huang2024}.

This study has several strengths. Our dataset is not in the public domain, ensuring it has never contributed to LLM's training data. This enhances the external validity of our results by testing the LLM on out-of-distribution data. The inclusion of human validation further strengthens our primary analysis by providing expert assessments of model outputs. However, there were limitations. The study was restricted to text-based questions, excluding multimodal questions involving images or tables. Advances in multimodal models could address this limitation in future research. Additionally, extensive domain expertise was required for dataset annotation and curation, presenting challenges in scalability. Finally, further validation across diverse data sources, such as electronic medical records, is necessary to ensure broader applicability and alignment with evolving clinical standards.

%% file: sections/conclusion.tex
\section{Conclusion}\label{conclusion}
This study supports the promise of SE as a tool for improving LLM model safety in clinical applications, suggesting a promising route for deploying LLMs in women's health. Future research should explore the development of domain-specific LLMs tailored to women's health, enabling more reliable responses and improved semantic understanding. As LLMs are introduced into healthcare settings, robust toolkits for auditing outputs and mitigating biases will be essential to maintain safety and efficacy.

%% file: sn-article.bbl

\begin{thebibliography}{0}
\ifx \bisbn   \undefined \def \bisbn  #1{ISBN #1}\fi
\ifx \binits  \undefined \def \binits#1{#1}\fi
\ifx \bauthor  \undefined \def \bauthor#1{#1}\fi
\ifx \batitle  \undefined \def \batitle#1{#1}\fi
\ifx \bjtitle  \undefined \def \bjtitle#1{#1}\fi
\ifx \bvolume  \undefined \def \bvolume#1{\textbf{#1}}\fi
\ifx \byear  \undefined \def \byear#1{#1}\fi
\ifx \bissue  \undefined \def \bissue#1{#1}\fi
\ifx \bfpage  \undefined \def \bfpage#1{#1}\fi
\ifx \blpage  \undefined \def \blpage #1{#1}\fi
\ifx \burl  \undefined \def \burl#1{\textsf{#1}}\fi
\ifx \doiurl  \undefined \def \doiurl#1{\url{https://doi.org/#1}}\fi
\ifx \betal  \undefined \def \betal{\textit{et al.}}\fi
\ifx \binstitute  \undefined \def \binstitute#1{#1}\fi
\ifx \binstitutionaled  \undefined \def \binstitutionaled#1{#1}\fi
\ifx \bctitle  \undefined \def \bctitle#1{#1}\fi
\ifx \beditor  \undefined \def \beditor#1{#1}\fi
\ifx \bpublisher  \undefined \def \bpublisher#1{#1}\fi
\ifx \bbtitle  \undefined \def \bbtitle#1{#1}\fi
\ifx \bedition  \undefined \def \bedition#1{#1}\fi
\ifx \bseriesno  \undefined \def \bseriesno#1{#1}\fi
\ifx \blocation  \undefined \def \blocation#1{#1}\fi
\ifx \bsertitle  \undefined \def \bsertitle#1{#1}\fi
\ifx \bsnm \undefined \def \bsnm#1{#1}\fi
\ifx \bsuffix \undefined \def \bsuffix#1{#1}\fi
\ifx \bparticle \undefined \def \bparticle#1{#1}\fi
\ifx \barticle \undefined \def \barticle#1{#1}\fi
\bibcommenthead
\ifx \bconfdate \undefined \def \bconfdate #1{#1}\fi
\ifx \botherref \undefined \def \botherref #1{#1}\fi
\ifx \url \undefined \def \url#1{\textsf{#1}}\fi
\ifx \bchapter \undefined \def \bchapter#1{#1}\fi
\ifx \bbook \undefined \def \bbook#1{#1}\fi
\ifx \bcomment \undefined \def \bcomment#1{#1}\fi
\ifx \oauthor \undefined \def \oauthor#1{#1}\fi
\ifx \citeauthoryear \undefined \def \citeauthoryear#1{#1}\fi
\ifx \endbibitem  \undefined \def \endbibitem {}\fi
\ifx \bconflocation  \undefined \def \bconflocation#1{#1}\fi
\ifx \arxivurl  \undefined \def \arxivurl#1{\textsf{#1}}\fi
\csname PreBibitemsHook\endcsname

\end{thebibliography}


\begin{thebibliography}{32}
    \ifx \bisbn   \undefined \def \bisbn  #1{ISBN #1}\fi
    \ifx \binits  \undefined \def \binits#1{#1}\fi
    \ifx \bauthor  \undefined \def \bauthor#1{#1}\fi
    \ifx \batitle  \undefined \def \batitle#1{#1}\fi
    \ifx \bjtitle  \undefined \def \bjtitle#1{#1}\fi
    \ifx \bvolume  \undefined \def \bvolume#1{\textbf{#1}}\fi
    \ifx \byear  \undefined \def \byear#1{#1}\fi
    \ifx \bissue  \undefined \def \bissue#1{#1}\fi
    \ifx \bfpage  \undefined \def \bfpage#1{#1}\fi
    \ifx \blpage  \undefined \def \blpage #1{#1}\fi
    \ifx \burl  \undefined \def \burl#1{\textsf{#1}}\fi
    \ifx \doiurl  \undefined \def \doiurl#1{\url{https://doi.org/#1}}\fi
    \ifx \betal  \undefined \def \betal{\textit{et al.}}\fi
    \ifx \binstitute  \undefined \def \binstitute#1{#1}\fi
    \ifx \binstitutionaled  \undefined \def \binstitutionaled#1{#1}\fi
    \ifx \bctitle  \undefined \def \bctitle#1{#1}\fi
    \ifx \beditor  \undefined \def \beditor#1{#1}\fi
    \ifx \bpublisher  \undefined \def \bpublisher#1{#1}\fi
    \ifx \bbtitle  \undefined \def \bbtitle#1{#1}\fi
    \ifx \bedition  \undefined \def \bedition#1{#1}\fi
    \ifx \bseriesno  \undefined \def \bseriesno#1{#1}\fi
    \ifx \blocation  \undefined \def \blocation#1{#1}\fi
    \ifx \bsertitle  \undefined \def \bsertitle#1{#1}\fi
    \ifx \bsnm \undefined \def \bsnm#1{#1}\fi
    \ifx \bsuffix \undefined \def \bsuffix#1{#1}\fi
    \ifx \bparticle \undefined \def \bparticle#1{#1}\fi
    \ifx \barticle \undefined \def \barticle#1{#1}\fi
    \bibcommenthead
    \ifx \bconfdate \undefined \def \bconfdate #1{#1}\fi
    \ifx \botherref \undefined \def \botherref #1{#1}\fi
    \ifx \url \undefined \def \url#1{\textsf{#1}}\fi
    \ifx \bchapter \undefined \def \bchapter#1{#1}\fi
    \ifx \bbook \undefined \def \bbook#1{#1}\fi
    \ifx \bcomment \undefined \def \bcomment#1{#1}\fi
    \ifx \oauthor \undefined \def \oauthor#1{#1}\fi
    \ifx \citeauthoryear \undefined \def \citeauthoryear#1{#1}\fi
    \ifx \endbibitem  \undefined \def \endbibitem {}\fi
    \ifx \bconflocation  \undefined \def \bconflocation#1{#1}\fi
    \ifx \arxivurl  \undefined \def \arxivurl#1{\textsf{#1}}\fi
    \csname PreBibitemsHook\endcsname
    
    \bibitem[\protect\citeauthoryear{Bachmann et~al.}{2024}]{Bachmann2024}
    \begin{botherref}
    \oauthor{\bsnm{Bachmann}, \binits{M.}},
    \oauthor{\bsnm{Duta}, \binits{I.}},
    \oauthor{\bsnm{Mazey}, \binits{E.}},
    \oauthor{\bsnm{Cooke}, \binits{W.}},
    \oauthor{\bsnm{Vatish}, \binits{M.}},
    \oauthor{\bsnm{Davis~Jones}, \binits{G.}}:
    Exploring the capabilities of chatgpt in women’s health: obstetrics and gynaecology.
    npj Women’s Health
    \textbf{2}(1)
    (2024)
    \doiurl{10.1038/s44294-024-00028-w}
    \end{botherref}
    \endbibitem
    
    \bibitem[\protect\citeauthoryear{Singhal}{2023}]{Singhal2023}
    \begin{barticle}
    \bauthor{\bsnm{Singhal}, \binits{K.e.a.}}:
    \batitle{Large language models encode clinical knowledge}.
    \bjtitle{Nature}
    \bvolume{620},
    \bfpage{172}--\blpage{180}
    (\byear{2023})
    \doiurl{10.1038/s41586-023-06210-3}
    \end{barticle}
    \endbibitem
    
    \bibitem[\protect\citeauthoryear{Gronowski and Yarbrough}{2018}]{Gronowski2018}
    \begin{barticle}
    \bauthor{\bsnm{Gronowski}, \binits{A.M.}},
    \bauthor{\bsnm{Yarbrough}, \binits{M.L.}}:
    \batitle{The women’s health diagnostic gap}.
    \bjtitle{Endocrinology}
    \bvolume{159},
    \bfpage{776}--\blpage{778}
    (\byear{2018})
    \doiurl{10.1210/en.2018-00468}
    \end{barticle}
    \endbibitem
    
    \bibitem[\protect\citeauthoryear{Clancy}{1992}]{Clancy1992}
    \begin{barticle}
    \bauthor{\bsnm{Clancy}, \binits{C.M.}}:
    \batitle{American women’s health care: A patchwork quilt with gaps}.
    \bjtitle{JAMA}
    \bvolume{268}(\bissue{14}),
    \bfpage{1918}
    (\byear{1992})
    \doiurl{10.1001/jama.1992.03490140126048}
    \end{barticle}
    \endbibitem
    
    \bibitem[\protect\citeauthoryear{Owens}{2008}]{Owens2008}
    \begin{barticle}
    \bauthor{\bsnm{Owens}, \binits{G.}}:
    \batitle{Gender differences in health care expenditures, resource utilization, and quality of care}.
    \bjtitle{Journal of Managed Care Pharmacy}
    \bvolume{14}(\bissue{3 Supp A}),
    \bfpage{2}--\blpage{6}
    (\byear{2008})
    \doiurl{10.18553/jmcp.2008.14.s3-a.2}
    \end{barticle}
    \endbibitem
    
    \bibitem[\protect\citeauthoryear{Amin}{2021}]{Amin2021}
    \begin{barticle}
    \bauthor{\bsnm{Amin}, \binits{A.e.a.}}:
    \batitle{Gender equality by 2045: reimagining a healthier future for women and girls}.
    \bjtitle{The Lancet}
    \bvolume{397},
    \bfpage{1276}--\blpage{1278}
    (\byear{2021})
    \doiurl{10.1016/S0140-6736(21)00712-6}
    \end{barticle}
    \endbibitem
    
    \bibitem[\protect\citeauthoryear{Peneva et~al.}{2015}]{Peneva2015}
    \begin{barticle}
    \bauthor{\bsnm{Peneva}, \binits{D.}},
    \bauthor{\bsnm{Xu}, \binits{X.}},
    \bauthor{\bsnm{Sutton}, \binits{A.}},
    \bauthor{\bsnm{Triche}, \binits{E.}},
    \bauthor{\bsnm{Ehrenkranz}, \binits{R.}},
    \bauthor{\bsnm{Paidas}, \binits{M.}},
    \bauthor{\bsnm{Stevens}, \binits{W.}},
    \bauthor{\bsnm{Shih}, \binits{T.}}:
    \batitle{The rising burden of preeclampsia in the united states impacts both maternal and child health}.
    \bjtitle{American Journal of Perinatology}
    \bvolume{33}(\bissue{04}),
    \bfpage{329}--\blpage{338}
    (\byear{2015})
    \doiurl{10.1055/s-0035-1564881}
    \end{barticle}
    \endbibitem
    
    \bibitem[\protect\citeauthoryear{Cascella et~al.}{2023}]{Cascella2023}
    \begin{botherref}
    \oauthor{\bsnm{Cascella}, \binits{M.}},
    \oauthor{\bsnm{Montomoli}, \binits{J.}},
    \oauthor{\bsnm{Bellini}, \binits{V.}},
    \oauthor{\bsnm{Bignami}, \binits{E.}}:
    Evaluating the feasibility of chatgpt in healthcare: An analysis of multiple clinical and research scenarios.
    Journal of Medical Systems
    \textbf{47}(1)
    (2023)
    \doiurl{10.1007/s10916-023-01925-4}
    \end{botherref}
    \endbibitem
    
    \bibitem[\protect\citeauthoryear{Antaki et~al.}{2023}]{Antaki2023}
    \begin{barticle}
    \bauthor{\bsnm{Antaki}, \binits{F.}},
    \bauthor{\bsnm{Touma}, \binits{S.}},
    \bauthor{\bsnm{Milad}, \binits{D.}},
    \bauthor{\bsnm{El-Khoury}, \binits{J.}},
    \bauthor{\bsnm{Duval}, \binits{R.}}:
    \batitle{Evaluating the performance of chatgpt in ophthalmology}.
    \bjtitle{Ophthalmology Science}
    \bvolume{3}(\bissue{4}),
    \bfpage{100324}
    (\byear{2023})
    \doiurl{10.1016/j.xops.2023.100324}
    \end{barticle}
    \endbibitem
    
    \bibitem[\protect\citeauthoryear{Wang et~al.}{2023}]{Wang2023}
    \begin{barticle}
    \bauthor{\bsnm{Wang}, \binits{C.}},
    \bauthor{\bsnm{Liu}, \binits{S.}},
    \bauthor{\bsnm{Yang}, \binits{H.}},
    \bauthor{\bsnm{Guo}, \binits{J.}},
    \bauthor{\bsnm{Wu}, \binits{Y.}},
    \bauthor{\bsnm{Liu}, \binits{J.}}:
    \batitle{Ethical considerations of using chatgpt in health care}.
    \bjtitle{Journal of Medical Internet Research}
    \bvolume{25},
    \bfpage{48009}
    (\byear{2023})
    \doiurl{10.2196/48009}
    \end{barticle}
    \endbibitem
    
    \bibitem[\protect\citeauthoryear{Kocoń et~al.}{2023}]{Kocon2023}
    \begin{barticle}
    \bauthor{\bsnm{Kocoń}, \binits{J.}},
    \bauthor{\bsnm{Cichecki}, \binits{I.}},
    \bauthor{\bsnm{Kaszyca}, \binits{O.}},
    \bauthor{\bsnm{Kochanek}, \binits{M.}},
    \bauthor{\bsnm{Szydło}, \binits{D.}},
    \bauthor{\bsnm{Baran}, \binits{J.}},
    \bauthor{\bsnm{Bielaniewicz}, \binits{J.}},
    \bauthor{\bsnm{Gruza}, \binits{M.}},
    \bauthor{\bsnm{Janz}, \binits{A.}},
    \bauthor{\bsnm{Kanclerz}, \binits{K.}},
    \bauthor{\bsnm{Kocoń}, \binits{A.}},
    \bauthor{\bsnm{Koptyra}, \binits{B.}},
    \bauthor{\bsnm{Mieleszczenko-Kowszewicz}, \binits{W.}},
    \bauthor{\bsnm{Miłkowski}, \binits{P.}},
    \bauthor{\bsnm{Oleksy}, \binits{M.}},
    \bauthor{\bsnm{Piasecki}, \binits{M.}},
    \bauthor{\bsnm{Radlinski}, \binits{L.}},
    \bauthor{\bsnm{Wojtasik}, \binits{K.}},
    \bauthor{\bsnm{Woźniak}, \binits{S.}},
    \bauthor{\bsnm{Kazienko}, \binits{P.}}:
    \batitle{Chatgpt: Jack of all trades, master of none}.
    \bjtitle{Information Fusion}
    \bvolume{99},
    \bfpage{101861}
    (\byear{2023})
    \doiurl{10.1016/j.inffus.2023.101861}
    \end{barticle}
    \endbibitem
    
    \bibitem[\protect\citeauthoryear{Temsah et~al.}{2023}]{Temsah2023}
    \begin{barticle}
    \bauthor{\bsnm{Temsah}, \binits{M.-H.}},
    \bauthor{\bsnm{Aljamaan}, \binits{F.}},
    \bauthor{\bsnm{Malki}, \binits{K.H.}},
    \bauthor{\bsnm{Alhasan}, \binits{K.}},
    \bauthor{\bsnm{Altamimi}, \binits{I.}},
    \bauthor{\bsnm{Aljarbou}, \binits{R.}},
    \bauthor{\bsnm{Bazuhair}, \binits{F.}},
    \bauthor{\bsnm{Alsubaihin}, \binits{A.}},
    \bauthor{\bsnm{Abdulmajeed}, \binits{N.}},
    \bauthor{\bsnm{Alshahrani}, \binits{F.S.}},
    \bauthor{\bsnm{Temsah}, \binits{R.}},
    \bauthor{\bsnm{Alshahrani}, \binits{T.}},
    \bauthor{\bsnm{Al-Eyadhy}, \binits{L.}},
    \bauthor{\bsnm{Alkhateeb}, \binits{S.M.}},
    \bauthor{\bsnm{Saddik}, \binits{B.}},
    \bauthor{\bsnm{Halwani}, \binits{R.}},
    \bauthor{\bsnm{Jamal}, \binits{A.}},
    \bauthor{\bsnm{Al-Tawfiq}, \binits{J.A.}},
    \bauthor{\bsnm{Al-Eyadhy}, \binits{A.}}:
    \batitle{Chatgpt and the future of digital health: A study on healthcare workers’ perceptions and expectations}.
    \bjtitle{Healthcare}
    \bvolume{11}(\bissue{13}),
    \bfpage{1812}
    (\byear{2023})
    \doiurl{10.3390/healthcare11131812}
    \end{barticle}
    \endbibitem
    
    \bibitem[\protect\citeauthoryear{Kung et~al.}{2023}]{Kung2023}
    \begin{barticle}
    \bauthor{\bsnm{Kung}, \binits{T.H.}},
    \bauthor{\bsnm{Cheatham}, \binits{M.}},
    \bauthor{\bsnm{Medenilla}, \binits{A.}},
    \bauthor{\bsnm{Sillos}, \binits{C.}},
    \bauthor{\bsnm{De~Leon}, \binits{L.}},
    \bauthor{\bsnm{Elepaño}, \binits{C.}},
    \bauthor{\bsnm{Madriaga}, \binits{M.}},
    \bauthor{\bsnm{Aggabao}, \binits{R.}},
    \bauthor{\bsnm{Diaz-Candido}, \binits{G.}},
    \bauthor{\bsnm{Maningo}, \binits{J.}},
    \bauthor{\bsnm{Tseng}, \binits{V.}}:
    \batitle{Performance of chatgpt on usmle: Potential for ai-assisted medical education using large language models}.
    \bjtitle{PLOS Digital Health}
    \bvolume{2}(\bissue{2}),
    \bfpage{0000198}
    (\byear{2023})
    \doiurl{10.1371/journal.pdig.0000198}
    \end{barticle}
    \endbibitem
    
    \bibitem[\protect\citeauthoryear{Kuhn et~al.}{2023}]{Kuhn2023}
    \begin{botherref}
    \oauthor{\bsnm{Kuhn}, \binits{L.}},
    \oauthor{\bsnm{Gal}, \binits{Y.}},
    \oauthor{\bsnm{Farquhar}, \binits{S.}}:
    Semantic uncertainty: Linguistic invariances for uncertainty estimation in natural language generation
    (2023)
    \doiurl{10.48550/ARXIV.2302.09664}
    {\href{https://arxiv.org/abs/2302.09664}{{arXiv:2302.09664}}}
    {[cs.CL]}
    \end{botherref}
    \endbibitem
    
    \bibitem[\protect\citeauthoryear{Farquhar et~al.}{2024}]{Farquhar2024}
    \begin{barticle}
    \bauthor{\bsnm{Farquhar}, \binits{S.}},
    \bauthor{\bsnm{Kossen}, \binits{J.}},
    \bauthor{\bsnm{Kuhn}, \binits{L.}},
    \bauthor{\bsnm{Gal}, \binits{Y.}}:
    \batitle{Detecting hallucinations in large language models using semantic entropy}.
    \bjtitle{Nature}
    \bvolume{630}(\bissue{8017}),
    \bfpage{625}--\blpage{630}
    (\byear{2024})
    \doiurl{10.1038/s41586-024-07421-0}
    \end{barticle}
    \endbibitem
    
    \bibitem[\protect\citeauthoryear{{OpenAI}}{}]{openAIAPI}
    \begin{botherref}
    \oauthor{\bsnm{{OpenAI}}}:
    {OpenAI API}.
    \url{https://openai.com/api/}.
    Accessed: 2024-09-30
    \end{botherref}
    \endbibitem
    
    \bibitem[\protect\citeauthoryear{{OpenAI}}{}]{OpenAIPrompt}
    \begin{botherref}
    \oauthor{\bsnm{{OpenAI}}}:
    {OpenAI} Prompt Engineering.
    \url{https://platform.openai.com/docs/guides/prompt-engineering}.
    Accessed: 2024-09-30
    \end{botherref}
    \endbibitem
    
    \bibitem[\protect\citeauthoryear{White et~al.}{2023}]{White2023}
    \begin{botherref}
    \oauthor{\bsnm{White}, \binits{J.}},
    \oauthor{\bsnm{Fu}, \binits{Q.}},
    \oauthor{\bsnm{Hays}, \binits{S.}},
    \oauthor{\bsnm{Sandborn}, \binits{M.}},
    \oauthor{\bsnm{Olea}, \binits{C.}},
    \oauthor{\bsnm{Gilbert}, \binits{H.}},
    \oauthor{\bsnm{Elnashar}, \binits{A.}},
    \oauthor{\bsnm{Spencer-Smith}, \binits{J.}},
    \oauthor{\bsnm{Schmidt}, \binits{D.C.}}:
    A prompt pattern catalog to enhance prompt engineering with chatgpt
    (2023)
    \doiurl{10.48550/ARXIV.2302.11382}
    {\href{https://arxiv.org/abs/2302.11382}{{arXiv:2302.11382}}}
    {[cs.SE]}
    \end{botherref}
    \endbibitem
    
    \bibitem[\protect\citeauthoryear{Padó et~al.}{2009}]{Pado2009}
    \begin{barticle}
    \bauthor{\bsnm{Padó}, \binits{S.}},
    \bauthor{\bsnm{Cer}, \binits{D.}},
    \bauthor{\bsnm{Galley}, \binits{M.}},
    \bauthor{\bsnm{Jurafsky}, \binits{D.}},
    \bauthor{\bsnm{Manning}, \binits{C.D.}}:
    \batitle{Measuring machine translation quality as semantic equivalence: A metric based on entailment features}.
    \bjtitle{Machine Translation}
    \bvolume{23}(\bissue{2–3}),
    \bfpage{181}--\blpage{193}
    (\byear{2009})
    \doiurl{10.1007/s10590-009-9060-y}
    \end{barticle}
    \endbibitem
    
    \bibitem[\protect\citeauthoryear{Penny-Dimri}{2025}]{codebaseRepo}
    \begin{botherref}
    \oauthor{\bsnm{Penny-Dimri}, \binits{J.C.}}:
    Reducing Large Language Model Safety Risks in Women's Health using Semantic Entropy.
    GitHub
    (2025).
    \doiurl{10.5281/zenodo.14933338} .
    \url{https://github.com/jahanpd/semantic_entropy_in_womens_health}
    \end{botherref}
    \endbibitem
    
    \bibitem[\protect\citeauthoryear{Barrault et~al.}{2024}]{lcm2024}
    \begin{botherref}
    \oauthor{\bsnm{Barrault}, \binits{L.}},
    \oauthor{\bsnm{Duquenne}, \binits{P.-A.}},
    \oauthor{\bsnm{Elbayad}, \binits{M.}},
    \oauthor{\bsnm{Kozhevnikov}, \binits{A.}},
    \oauthor{\bsnm{Alastruey}, \binits{B.}},
    \oauthor{\bsnm{Andrews}, \binits{P.}},
    \oauthor{\bsnm{Coria}, \binits{M.}},
    \oauthor{\bsnm{Couairon}, \binits{G.}},
    \oauthor{\bsnm{Costa-jussà}, \binits{M.R.}},
    \oauthor{\bsnm{Dale}, \binits{D.}},
    \oauthor{\bsnm{Elsahar}, \binits{H.}},
    \oauthor{\bsnm{Heffernan}, \binits{K.}},
    \oauthor{\bsnm{Janeiro}, \binits{J.M.}},
    \oauthor{\bsnm{Tran}, \binits{T.}},
    \oauthor{\bsnm{Ropers}, \binits{C.}},
    \oauthor{\bsnm{Sánchez}, \binits{E.}},
    \oauthor{\bsnm{Roman}, \binits{R.S.}},
    \oauthor{\bsnm{Mourachko}, \binits{A.}},
    \oauthor{\bsnm{Saleem}, \binits{S.}},
    \oauthor{\bsnm{Schwenk}, \binits{H.}}:
    Large Concept Models: Language Modeling in a Sentence Representation Space
    (2024).
    \url{https://arxiv.org/abs/2412.08821}
    \end{botherref}
    \endbibitem
    
    \bibitem[\protect\citeauthoryear{Júnior and Vitorino}{2024}]{Junior2024}
    \begin{botherref}
    \oauthor{\bsnm{Júnior}, \binits{G.H.Y.}},
    \oauthor{\bsnm{Vitorino}, \binits{L.M.}}:
    Large language models in healthcare: An urgent call for ongoing, rigorous validation.
    Journal of Medical Systems
    \textbf{48}(1)
    (2024)
    \doiurl{10.1007/s10916-024-02126-3}
    \end{botherref}
    \endbibitem
    
    \bibitem[\protect\citeauthoryear{Bedi et~al.}{2024}]{Bedi2024}
    \begin{barticle}
    \bauthor{\bsnm{Bedi}, \binits{S.}},
    \bauthor{\bsnm{Jain}, \binits{S.S.}},
    \bauthor{\bsnm{Shah}, \binits{N.H.}}:
    \batitle{Evaluating the clinical benefits of llms}.
    \bjtitle{Nature Medicine}
    \bvolume{30}(\bissue{9}),
    \bfpage{2409}--\blpage{2410}
    (\byear{2024})
    \doiurl{10.1038/s41591-024-03181-6}
    \end{barticle}
    \endbibitem
    
    \bibitem[\protect\citeauthoryear{Salvagno et~al.}{2023}]{Salvagno2023}
    \begin{botherref}
    \oauthor{\bsnm{Salvagno}, \binits{M.}},
    \oauthor{\bsnm{Taccone}, \binits{F.S.}},
    \oauthor{\bsnm{Gerli}, \binits{A.G.}}:
    Can artificial intelligence help for scientific writing?
    Critical Care
    \textbf{27}(1)
    (2023)
    \doiurl{10.1186/s13054-023-04380-2}
    \end{botherref}
    \endbibitem
    
    \bibitem[\protect\citeauthoryear{Pal et~al.}{2023}]{Pal2023}
    \begin{botherref}
    \oauthor{\bsnm{Pal}, \binits{A.}},
    \oauthor{\bsnm{Umapathi}, \binits{L.K.}},
    \oauthor{\bsnm{Sankarasubbu}, \binits{M.}}:
    Med-halt: Medical domain hallucination test for large language models
    (2023)
    \doiurl{10.48550/ARXIV.2307.15343}
    {\href{https://arxiv.org/abs/2307.15343}{{arXiv:2307.15343}}}
    {[cs.CL]}
    \end{botherref}
    \endbibitem
    
    \bibitem[\protect\citeauthoryear{Eoh et~al.}{2024}]{Eoh2024}
    \begin{barticle}
    \bauthor{\bsnm{Eoh}, \binits{K.J.}},
    \bauthor{\bsnm{Kwon}, \binits{G.Y.}},
    \bauthor{\bsnm{Lee}, \binits{E.J.}},
    \bauthor{\bsnm{Lee}, \binits{J.}},
    \bauthor{\bsnm{Lee}, \binits{I.}},
    \bauthor{\bsnm{Kim}, \binits{Y.T.}},
    \bauthor{\bsnm{Nam}, \binits{E.J.}}:
    \batitle{Efficacy of large language models and their potential in obstetrics and gynecology education}.
    \bjtitle{Obstetrics \&; Gynecology Science}
    \bvolume{67}(\bissue{6}),
    \bfpage{550}--\blpage{556}
    (\byear{2024})
    \doiurl{10.5468/ogs.24211}
    \end{barticle}
    \endbibitem
    
    \bibitem[\protect\citeauthoryear{Sengupta et~al.}{2023}]{Sengupta2023}
    \begin{barticle}
    \bauthor{\bsnm{Sengupta}, \binits{P.}},
    \bauthor{\bsnm{Dutta}, \binits{S.}},
    \bauthor{\bsnm{Chakravarthi}, \binits{S.}},
    \bauthor{\bsnm{Jegasothy}, \binits{R.}},
    \bauthor{\bsnm{Jeganathan}, \binits{R.}},
    \bauthor{\bsnm{Pichumani}, \binits{A.}}:
    \batitle{Comparative efficacy of chatgpt 3.5, chatgpt 4, and other large language models in gynecology and infertility research}.
    \bjtitle{Gynecology and Obstetrics Clinical Medicine}
    \bvolume{3}(\bissue{4}),
    \bfpage{203}--\blpage{206}
    (\byear{2023})
    \doiurl{10.1016/j.gocm.2023.09.002}
    \end{barticle}
    \endbibitem
    
    \bibitem[\protect\citeauthoryear{Chen et~al.}{2022}]{Chen2022}
    \begin{botherref}
    \oauthor{\bsnm{Chen}, \binits{H.-T.}},
    \oauthor{\bsnm{Zhang}, \binits{M.J.Q.}},
    \oauthor{\bsnm{Choi}, \binits{E.}}:
    Rich Knowledge Sources Bring Complex Knowledge Conflicts: Recalibrating Models to Reflect Conflicting Evidence
    (2022).
    \url{https://arxiv.org/abs/2210.13701}
    \end{botherref}
    \endbibitem
    
    \bibitem[\protect\citeauthoryear{Levy et~al.}{2024}]{Levy2024}
    \begin{botherref}
    \oauthor{\bsnm{Levy}, \binits{S.}},
    \oauthor{\bsnm{Karver}, \binits{T.S.}},
    \oauthor{\bsnm{Adler}, \binits{W.D.}},
    \oauthor{\bsnm{Kaufman}, \binits{M.R.}},
    \oauthor{\bsnm{Dredze}, \binits{M.}}:
    Evaluating biases in context-dependent health questions
    (2024)
    \doiurl{10.48550/ARXIV.2403.04858}
    {\href{https://arxiv.org/abs/2403.04858}{{arXiv:2403.04858}}}
    {[cs.CL]}
    \end{botherref}
    \endbibitem
    
    \bibitem[\protect\citeauthoryear{Shankar et~al.}{2024}]{Shankar2024}
    \begin{botherref}
    \oauthor{\bsnm{Shankar}, \binits{S.}},
    \oauthor{\bsnm{Zamfirescu-Pereira}, \binits{J.D.}},
    \oauthor{\bsnm{Hartmann}, \binits{B.}},
    \oauthor{\bsnm{Parameswaran}, \binits{A.G.}},
    \oauthor{\bsnm{Arawjo}, \binits{I.}}:
    Who validates the validators? aligning llm-assisted evaluation of llm outputs with human preferences
    (2024)
    \doiurl{10.48550/ARXIV.2404.12272}
    {\href{https://arxiv.org/abs/2404.12272}{{arXiv:2404.12272}}}
    {[cs.HC]}
    \end{botherref}
    \endbibitem
    
    \bibitem[\protect\citeauthoryear{Ziegler et~al.}{2019}]{Ziegler2019}
    \begin{botherref}
    \oauthor{\bsnm{Ziegler}, \binits{D.M.}},
    \oauthor{\bsnm{Stiennon}, \binits{N.}},
    \oauthor{\bsnm{Wu}, \binits{J.}},
    \oauthor{\bsnm{Brown}, \binits{T.B.}},
    \oauthor{\bsnm{Radford}, \binits{A.}},
    \oauthor{\bsnm{Amodei}, \binits{D.}},
    \oauthor{\bsnm{Christiano}, \binits{P.}},
    \oauthor{\bsnm{Irving}, \binits{G.}}:
    Fine-tuning language models from human preferences
    (2019)
    \doiurl{10.48550/ARXIV.1909.08593}
    {\href{https://arxiv.org/abs/1909.08593}{{arXiv:1909.08593}}}
    {[cs.CL]}
    \end{botherref}
    \endbibitem
    
    \bibitem[\protect\citeauthoryear{Huang et~al.}{2024}]{Huang2024}
    \begin{botherref}
    \oauthor{\bsnm{Huang}, \binits{X.}},
    \oauthor{\bsnm{Ruan}, \binits{W.}},
    \oauthor{\bsnm{Huang}, \binits{W.}},
    \oauthor{\bsnm{Jin}, \binits{G.}},
    \oauthor{\bsnm{Dong}, \binits{Y.}},
    \oauthor{\bsnm{Wu}, \binits{C.}},
    \oauthor{\bsnm{Bensalem}, \binits{S.}},
    \oauthor{\bsnm{Mu}, \binits{R.}},
    \oauthor{\bsnm{Qi}, \binits{Y.}},
    \oauthor{\bsnm{Zhao}, \binits{X.}},
    \oauthor{\bsnm{Cai}, \binits{K.}},
    \oauthor{\bsnm{Zhang}, \binits{Y.}},
    \oauthor{\bsnm{Wu}, \binits{S.}},
    \oauthor{\bsnm{Xu}, \binits{P.}},
    \oauthor{\bsnm{Wu}, \binits{D.}},
    \oauthor{\bsnm{Freitas}, \binits{A.}},
    \oauthor{\bsnm{Mustafa}, \binits{M.A.}}:
    A survey of safety and trustworthiness of large language models through the lens of verification and validation.
    Artificial Intelligence Review
    \textbf{57}(7)
    (2024)
    \doiurl{10.1007/s10462-024-10824-0}
    \end{botherref}
    \endbibitem
    
    \end{thebibliography}
